%% file: main.tex
\documentclass{article}
\PassOptionsToPackage{numbers,compress}{natbib}

\usepackage[final]{neurips_2021}

\usepackage[dvipsnames]{xcolor}         
\definecolor{linkColor}{rgb}{0.2,0.4,0.6}
\usepackage[utf8]{inputenc} %
\usepackage[T1]{fontenc}    %
\usepackage[colorlinks=true,linkcolor=linkColor,citecolor=linkColor,filecolor=linkColor,urlcolor=linkColor]{hyperref}       %
\usepackage{url}            %
\usepackage{booktabs}       %
\usepackage{amsfonts}       %
\usepackage{nicefrac}       %
\usepackage{microtype}      %
\usepackage{xcolor}         %
\usepackage{graphicx}
\usepackage{longtable}
\usepackage{caption}
\usepackage{mdframed}
\usepackage{subcaption}
\usepackage{multirow}
\usepackage{placeins}
\usepackage{multicol}
\usepackage{makecell}
\usepackage[normalem]{ulem} %
\usepackage{wrapfig}
\usepackage[percent]{overpic}
\usepackage{lipsum}
\usepackage{csquotes}
\usepackage[OT2,T1]{fontenc}
\usepackage[english]{babel}
\usepackage{tablefootnote}
\usepackage{pdfpages}
\usepackage{listings}
\usepackage{float}
\usepackage{tabularx}
\usepackage{longtable} 
\usepackage{macros}
\usepackage{tocloft}
\usepackage{xltabular}
\usepackage[skins]{tcolorbox}
\tcbuselibrary{breakable}
\usepackage{amsmath}

\newmdenv[
  font=\ttfamily\small,
  linewidth=0.5pt,
  innerleftmargin=10pt,
  innerrightmargin=10pt,
  innertopmargin=10pt,
  innerbottommargin=10pt,
]{monobox}

\newif\ifcomment

\newcommand\ours{\textsc{RedStone}}

\title{%
    \ours{}\raisebox{-0.13\height}{\includegraphics[height=2.0ex]{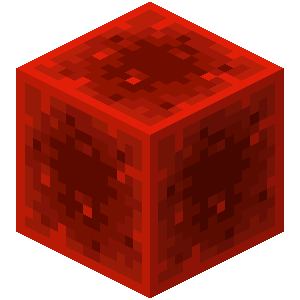}}: Curating General, Code, Math, and QA Data for Large Language Models
}  

\author{Yaoyao CHANG, Lei CUI, Li DONG, Shaohan HUANG, Yangyu HUANG, \\
\bf Yupan HUANG, Scarlett LI, Tengchao LV, Shuming MA, Qinzheng SUN, \\
\bf Wenhui WANG, Furu WEI, Ying XIN, Mao YANG, Qiufeng YIN, Xingxing ZHANG.
\\\\
{\bf Microsoft Research}\thanks{The contributors are listed in the alphabetical order by last names. Corresponding author: Furu WEI (\href{mailto:fuwei@microsoft.com}{fuwei@microsoft.com}.)}\\
\url{https://aka.ms/GeneralAI} \\
\\
Project Page: \url{https://aka.ms/redstone}}

\begin{document}
\commenttrue %

\maketitle
\input{chapters/abstract}
\newpage



\input{chapters/introduction}


\input{chapters/redstone}

\input{chapters/experiments}

\input{chapters/related_works}
\input{chapters/conclusion}

\newpage
\bibliography{references}{}
\bibliographystyle{alpha}

\newpage
\input{chapters/appendix}

\end{document}

%% file: chapters/abstract.tex
\begin{abstract}

Pre-training Large Language Models (LLMs) on high-quality, meticulously curated datasets is widely recognized as critical for enhancing their performance and generalization capabilities. 
This study explores the untapped potential of Common Crawl as a comprehensive and flexible resource for pre-training LLMs, addressing both general-purpose language understanding and specialized domain knowledge. We introduce \textbf{\ours{}}, an innovative and scalable pipeline engineered to extract and process data from Common Crawl, facilitating the creation of extensive and varied pre-training datasets. Unlike traditional datasets, which often require expensive curation and domain-specific expertise, \ours{} leverages the breadth of Common Crawl to deliver datasets tailored to a wide array of domains. In this work, we exemplify its capability by constructing pre-training datasets across multiple fields, including general language understanding, code, mathematics, and question-answering tasks. The flexibility of \ours{} allows for easy adaptation to other specialized domains, significantly lowering the barrier to creating valuable domain-specific datasets. Our findings demonstrate that Common Crawl, when harnessed through effective pipelines like \ours{}, can serve as a rich, renewable source of pre-training data, unlocking new avenues for domain adaptation and knowledge discovery in LLMs. This work also underscores the importance of innovative data acquisition strategies and highlights the role of web-scale data as a powerful resource in the continued evolution of LLMs. RedStone code and data samples will be publicly available at \url{https://aka.ms/redstone}.

\end{abstract}

%% file: chapters/introduction.tex
\section{Introduction}
Large Language Models (LLMs) have demonstrated remarkable potential as highly capable AI assistants, particularly in complex reasoning tasks that require expert knowledge across diverse fields~\cite{radford2019language, brown2020language, touvron2023llama, dubey2024llama}. The recent advancements in LLMs have been driven not only by increasing model sizes but also by scaling up dataset sizes correspondingly. These advancements have created a demand for robust data curation pipelines capable of mining vast, diverse datasets across various domains~\cite{li2023textbooks,brown2020language,chowdhery2023palm,touvron2023llama,dubey2024llama}.

In addition to general domain data that spans a wide range of topics, there is an increasing need for domain-specific data tailored to specialized fields such as mathematics and code. These domains require high-quality, domain-specific knowledge to enhance model performance. Current approaches for constructing domain-specific datasets often rely on proprietary resources~\cite{kocetkov2022stack}, synthetic LLM-generated data~\cite{xu2023wizardlm, lewkowycz2022solving}, or manually annotated data~\cite{brown2020language, llama3modelcard, textbooks2}, which is often time-consuming and labor-intensive. Moreover, these sources are limited in scope and scale, often resulting in small dataset sizes.
Common Crawl, a massive open-access web archive, offers a wealth of diverse, high-quality data across various domains. Although recent efforts, such as RefinedWeb~\cite{penedo2023refinedweb} and Redpajama v2~\cite{together2023redpajama}, have explored mining data from Common Crawl, their focus has primarily been on general data extraction, with limited attention to domain-specific data.

In this paper, we introduce \textbf{\ours{}}, a comprehensive pipeline designed to efficiently extract and filter large-scale datasets from Common Crawl, applicable to both {general domain data} and {domain-specific data}.
As shown in Figure~\ref{figure:data_constructed_with_RedStone}, we propose using Common Crawl as a valuable source for mining knowledge, as its web data often includes annotations that provide extensive context and reasoning. 
For instance, code snippets and mathematics equations in Common Crawl are frequently accompanied by discussions, explanations, and even execution results, offering richer insights than standalone source code and equations.
Open-domain question answering is often interleaved with other text content in the document, which may include reading passages, explanations, or other informative content.

The \ours{} pipeline defines general or domain-specific data formats, then extracts and filters relevant data from Common Crawl. The Extraction module processes raw web data using techniques like pattern recognition and NLP to capture the necessary formats for training. The Filtering module refines this data, retaining only the most relevant content through keyword searches, regular expressions, and machine learning models. This approach efficiently generates large-scale datasets, streamlining the process of data sourcing, synthesis, and annotation.
For example, in the process of obtaining open question answering data, the extraction module retrieves the main content from web pages, using WET files and CCNet's paragraph-level deduplication method to ensure completeness. The filtering module then employs a two-stage approach: rule-based filtering identifies common patterns for questions and answers using keywords, while model-based filtering applies a self-trained classifier to further refine the data. This process results in a large, high-quality dataset of open-domain questions, efficiently combining rule-based and machine learning techniques.

We curate general domain data from two available formats of Common Crawl, resulting in the \textbf{\ours{}-Web} dataset, which contains approximately 3.17 trillion tokens of general knowledge. For domain-specific data, we focus on constructing datasets for code, mathematics, and question answering, as these areas are crucial to the core capabilities of large language models. The resulting datasets—\textbf{\ours{}-Code}, \textbf{\ours{}-Math}, and \textbf{\ours{}-QA}—comprise 250.2, 15.9, and 51.4 billion tokens, respectively.

Our constructed general domain data, sourced from diverse web pages, enhances the model's understanding of language in a broad context, while domain-specific data provides specialized knowledge in fields such as code, mathematics, and question answering.
We evaluated \ours{} across a range of tasks, demonstrating its effectiveness in enhancing model performance. \ours{}-Web outperformed other open-source datasets in common sense reasoning tasks, such as ARC-e, HellaSwag, OpenBookQA, and PIQA. Incorporating \ours{}-Code into the general dataset significantly boosted performance in code generation benchmarks, including HumanEval and MBPP. Similarly, \ours{}-Math outperformed existing datasets on mathematics benchmarks like GSM8k and MATH, demonstrating improved perplexity. Finally, \ours{}-QA achieved the highest scores in question-answering tasks, particularly on the MMLU benchmark, confirming the strength of our pipeline in extracting high-quality data across diverse domains.

Our contributions are summarized as follows:

\begin{itemize}
\item We introduce \ours{}, a pipeline for obtaining large-scale diverse data from the web with a detailed description of the pipeline's components and workflow for reproducibility.

\item Using \ours{}, we mined and constructed several large-scale datasets totaling 3.48 trillion tokens, encompassing both general domain and domain-specific data. The {\ours{}-Web} dataset contains approximately 3.17 trillion tokens of general knowledge, while {\ours{}-Code}, {\ours{}-Math}, and {\ours{}-QA} consist of 250.2 billion, 15.9 billion, and 51.4 billion tokens, respectively.

\item We demonstrate the effectiveness of \ours{} through extensive evaluations across common sense reasoning, code generation, and mathematics tasks. Our general domain dataset, \ours{}-Web, outperforms existing open-source datasets in common sense reasoning benchmarks, while the inclusion of \ours{}-Code and \ours{}-Math significantly improves model performance in code generation and mathematical problem solving. Additionally, \ours{}-QA shows notable advancements in question-answering tasks, further establishing \ours{} as a robust dataset for diverse pre-training applications.

\item We discuss strategies for maximizing the potential of Common Crawl. To the best of our knowledge, this is the first systematic exploration of the full potential of Common Crawl, demonstrating its capability as a rich resource for various domain-specific contents.

\end{itemize}

%% file: chapters/redstone.tex
\section{\ours{}}

\ours{} is a pipeline designed for the large-scale extraction of various data from web data consisting of \textbf{Extraction} and \textbf{Filtering} modules as detailed below. 

\begin{itemize}

\item \textbf{Extraction.} 
Extracting raw data to obtain the required format for training. Extraction can involve the use of pattern recognition, natural language processing, and other computational methods to obtain the desired information.

\item \textbf{Filtering.} 
Selecting relevant data and excluding unnecessary or irrelevant information to focus on the most pertinent data for analysis. Filtering techniques can include keyword searches, regular expressions, and machine learning models to ensure that only the most relevant data is retained.

\end{itemize}

This section will detail the process by which \ours{} employs these two modules to construct various types of data. Additionally, in terms of dataset types, \ours{} categorizes data into general domain data and domain-specific data, below are the specific definitions of these two types.

\begin{itemize}
\item \textbf{General domain data} helps the model understand language in a broad context, enhancing its comprehension of the real world. Currently, the primary source of pre-training data for LLMs is general domain data, which is crucial for their basic performance. However, existing pipelines for data processing still have limitations, to improve efficiency, certain steps have been simplified. \ours{} integrates current mainstream pipelines, restructures the processing steps, and refines certain stages to produce a high-quality general domain dataset, \ours{}-Web.

\item \textbf{Domain-specific data} refers to detailed, context-specific information pertinent to particular fields, tasks, or scenarios, such as mathematics, coding, reasoning, medicine, law, engineering, and finance. 
This high-quality data can significantly improve model performance but is often time-consuming and labor-intensive to construct.
Common Crawl contains a substantial amount of domain-specific data, which can be extracted at scale using specialized pipelines and filtering techniques to enhance quality.
Additionally, web data often includes discussions and explanations, which help LLMs gain a deeper understanding of specific knowledge.
Taking code, math, and QA as examples, \ours{} has constructed large-scale domain-specific datasets from Common Crawl, namely \ours{}-Code, \ours{}-Math, and \ours{}-QA.

\end{itemize}

\begin{figure}[H]
\begin{center}
\includegraphics[width=\textwidth]{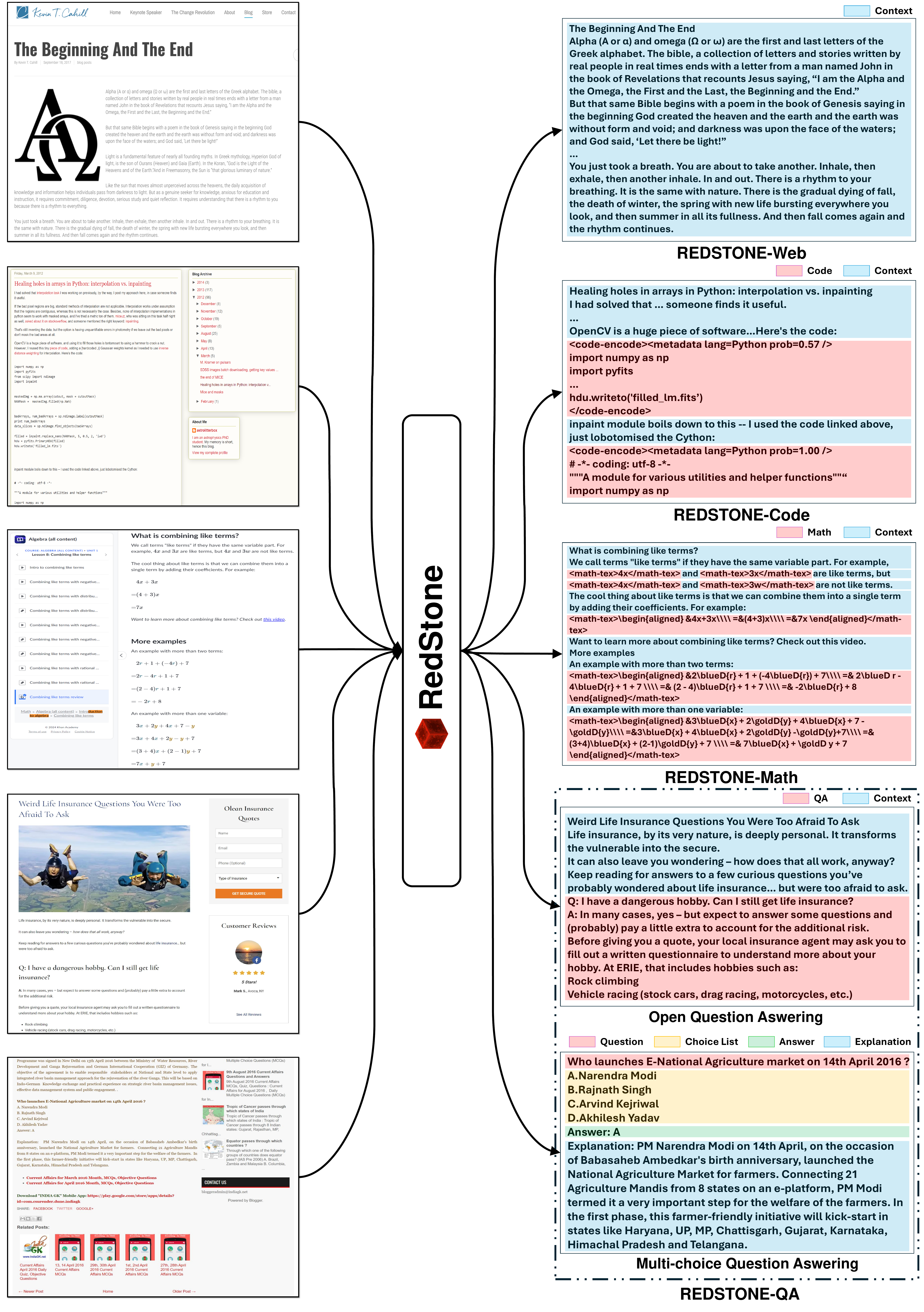}
\end{center}
   \caption{Using \ours{}, we created two types of data: \textbf{general domain data} and \textbf{domain-specific data}. General domain data comprises \ours{}-Web, it does not specify a data domain, allowing the model to learn common knowledge across various domains. Domain-specific data includes \ours{}-Code, \ours{}-Math, and \ours{}-QA, enabling the model to acquire specialized knowledge in particular areas or formats. \textbf{Each example type features the original webpage screenshot on the left and the corresponding data processed by \ours{} on the right.}}
\label{figure:data_constructed_with_RedStone}
\end{figure}

The construction processes for the two types of data will be introduced separately. Details on \ours{}-Web within general domain data are provided in Section~\ref{sec:data_web}, while \ours{}-Code, \ours{}-Math, and \ours{}-QA within domain-specific data are described in Section~\ref{sec:data_code},~\ref{sec:data_math}, and~\ref{sec:data_qa}, respectively.

\begin{table}[t]
\centering
\setlength{\tabcolsep}{5pt}
\begin{tabular}{crr}
\toprule
& \bf Dataset & \bf Tokens (B) \\
\midrule
General Domain Data & \ours{}-Web & 3,170.2 \\
\midrule
\multirow{3}{*}{Domain-specific Data}
& \ours{}-Code & 250.2 \\
& \ours{}-Math & 15.9 \\
& \ours{}-QA & 51.4 \\
\bottomrule
\end{tabular}
\caption{Dataset statistics for general and specific domain data constructed using \ours{}.}
\label{tab:dataset}
\end{table}

\subsection{\ours{}-Web} 
\label{sec:data_web}

\begin{figure}[h]
\begin{center}
\includegraphics[width=5.0in]{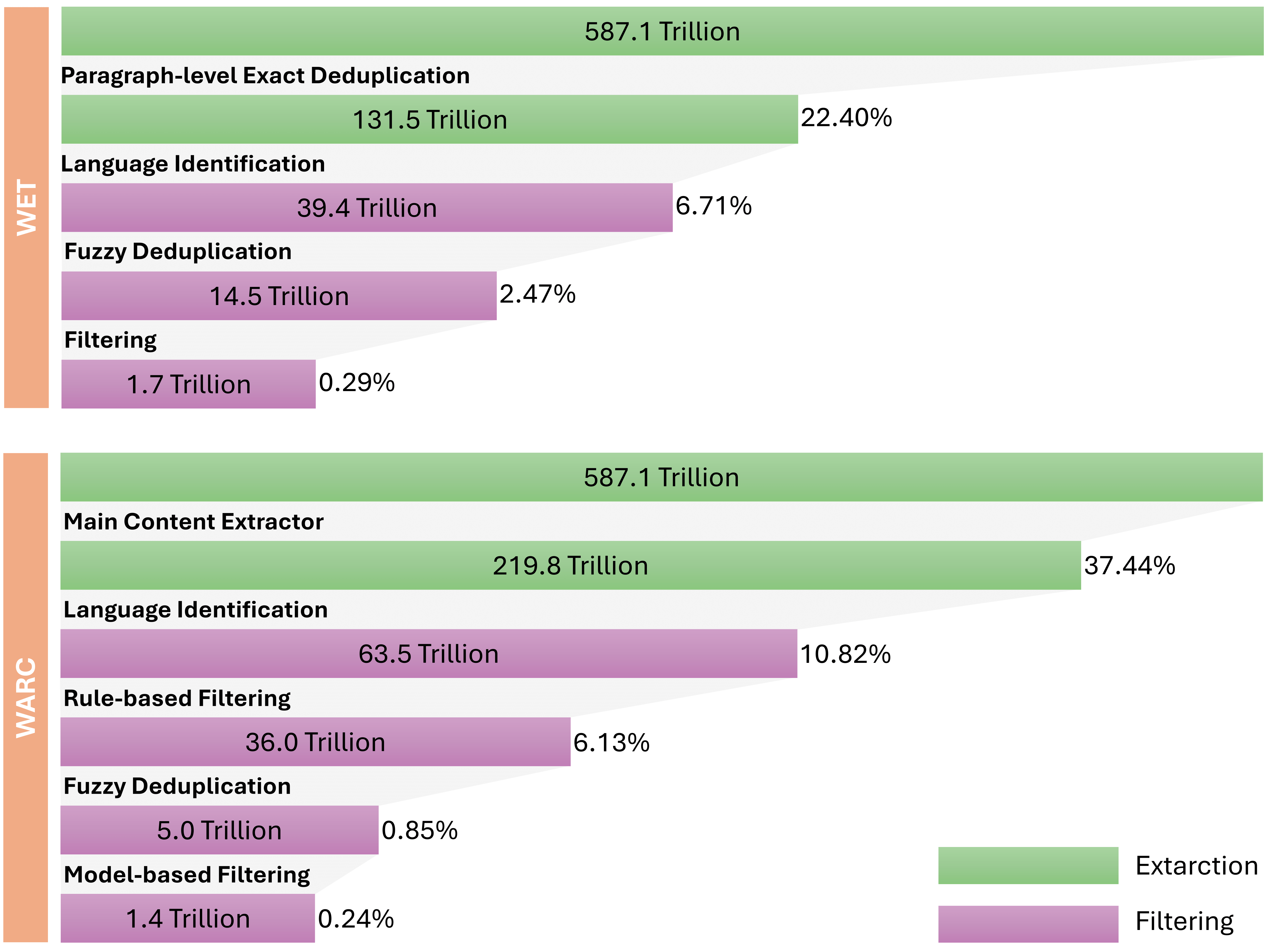}
\end{center}
   \caption{Subsequent stages of \ours{}-Web. \ours{} processes Common Crawl data in separate steps, handling WARC and WET files independently before merging them to increase the token count. Over 99\% of the tokens in Common Crawl are removed during processing. Since WARC files are in HTML format and inconvenient for token counting, and WARC and WET files represent different forms of the same data, the token count from WET files is used as the original token count for both formats.}
\label{figure:subsequent_general}
\end{figure}

To construct general domain data, we extract the main content from raw web pages and filter data based on the quality of the pages.
As illustrated in Figure~\ref{figure:subsequent_general} and shown in Table~\ref{tab:general_data_stat}, through extraction and filtering, we obtained a comprehensive dataset, \textbf{\ours{}-Web}, containing approximately 3.17 trillion tokens of general knowledge.

\begin{table}[htbp]
\centering
\begin{tabular}{lccc}
\hline
\bf General Domain Data & \bf Tokens (T) & \bf Pages (M) & \bf Tokens per Page \\
\hline
WET & 1.74 & 856 & 2,032 \\
WARC & 1.43 & 1,223 & 1,169 \\
\hline
Total & 3.17 & 2,079 & 1,524 \\
\hline
\end{tabular}
\caption{\ours{}-Web from Common Crawl.}
\label{tab:general_data_stat}
\end{table}

\subsubsection{Extraction}
\label{sec:general_data_extraction}

Common Crawl is available in WARC and WET formats. WARC files contain the raw HTML responses, while WET files are pre-processed to include only plain text. Each record corresponds to a single webpage from a specific URL, representing an independent document or sample. 

We extract general domain data from Common Crawl in both WET and WARC formats, encompassing 93 CommonCrawl snapshots spanning from 2013-20 to 2023-40, resulting in two complementary datasets. Although these formats originate from the same data source and represent different forms of the same dataset, the differences in text extraction between WET and WARC lead to variations in the resulting plain text.

\paragraph{Text Extraction of WET.}
For text extraction from WET files, we adapt the deduplication strategy used in CCNet~\cite{wenzek2019ccnet} to eliminate noisy text from web pages.
Specifically, the deduplication strategy removes frequently appeared common segments (e.g., 'sign in,' 'follow,' 'about') on each page by identifying the duplicates against other paragraph within a shard of files. Instead of limiting the deduplication process to 5GB segments, we expanded the search to one snapshot of the Common Crawl.
This comprehensive approach enhances the ability to detect and remove noisy text, resulting in cleaner, higher-quality extracted content.

\paragraph{Text Extraction of WARC.}
For text extraction from WARC files, we followed the implementation used by RefinedWeb~\cite{penedo2023refinedweb}, utilizing the Trafilatura~\cite{barbaresi-2021-trafilatura}. Trafilatura is designed to extract meaningful content from web pages while excluding irrelevant sections such as menus, headers, and footers. It employs a combination of techniques including HTML parsing, content density analysis, and feature extraction to identify and isolate the main body of text.

Once the high-density text regions are identified, Trafilatura applies feature extraction techniques to further refine the selection. This includes looking for patterns and structures typical of main content, such as paragraphs, headings, and continuous text blocks, while ignoring sections with numerous links or short, unrelated text fragments.

By employing these techniques, Trafilatura effectively filters out non-essential parts of the web pages, resulting in cleaner and more relevant text extraction from WARC files. This approach ensures that the extracted dataset is of higher quality, focusing on the core content while minimizing noise.

\subsubsection{Filtering}
Common Crawl contains a large number of low-quality pages, with RefinedWeb removing nearly 90\% of the documents originally in Common Crawl. Therefore, the effectiveness of document quality filtering significantly impacts the final quality of the dataset. High-quality datasets enable faster convergence and better model performance~\cite{together2023redpajama,lozhkov2024fineweb-edu,li2024datacomp}. Consequently, we employed a filtering module to eliminate low-quality texts. This filtering module consists of four stages: language filtering, rule-based filtering, model-based filtering, and deduplication.

\paragraph{Language Filtering.}
Common Crawl encompasses nearly all languages globally. We use fastText~\cite{joulin2016fasttext} as the language identification tool to filter out non-English pages with a confidence threshold of 0.5.

\paragraph{Rule-based Filtering.} Given the vast number of original pages, we initially use rule-based filtering to quickly sift through all pages, effectively reducing the subsequent filtering workload. 

For the main content extracted from WET files, we adhere to the filtering rules provided by CCNet, using length as a criterion with a threshold 300 to filter out short texts. 

For the main content extracted from WARC files, we follow the rules used by RefinedWeb, conducting repetition removal, document-level removal, and sentence-level removal. Repetition removal detects whether there is a significant amount of internal repetition within the text (e.g., a sentence repeated multiple times on a page), document-level removal determines whether to discard the current document based on its features, and sentence-level removal assesses whether to discard specific sentences based on their characteristics. 

Although the same set of processing rules can be applied to both WET and WARC files, we follow each pipeline's specific rules to filter separately. This makes the data more diverse, allowing the two datasets to complement each other and combine into a large-scale pre-training dataset. The detailed steps for WARC files are as follows.

\begin{itemize}
\item \textbf{Repetition Removal Rules. }
Documents that match any of following rules will be discarded.

\begin{itemize}
\item If the ratio of the number of duplicated sentences to the total sentence count exceeds 0.3;
\item If the ratio of the count of characters in duplicated sentences to the total character count exceeds 0.2;
\item If the ratio of the number of duplicated paragraphs to the total sentence count exceeds 0.3;
\item If the ratio of the count of characters in duplicated paragraphs to the total character count exceeds 0.2;
\item
for each \(n \in \{2...4\}\), if the ratio of the count of characters in most common n-gram to the total character count exceeds threshold, the detail thresholds are given in Table~\ref{table:ngram_rule_threshold}.
\item
for each \(n \in \{5...10\}\), if the ratio of the count of characters in duplicated n-grams (each character is only counted once regardless of its occurrence in several overlapping n-grams) to the total character count exceeds threshold, the detail thresholds are given in Table~\ref{table:ngram_rule_threshold}.
\end{itemize}

\begin{table}[hbtp]
\centering
\setlength{\tabcolsep}{10pt}
\begin{tabular}{lr}
\hline
\textbf{Rule} & \textbf{Threshold} \\
\hline
most common 2-gram & 0.20\\
most common 3-gram & 0.18\\
most common 4-gram & 0.16\\
duplicated 5-gram & 0.15\\
duplicated 6-gram & 0.14\\
duplicated 7-gram & 0.13\\
duplicated 8-gram & 0.12\\
duplicated 9-gram & 0.11\\
duplicated 10-gram & 0.10\\
\hline
\end{tabular}
\caption{Thresholds of n-gram character ratio related rules.}
\label{table:ngram_rule_threshold}
\end{table}

\item \textbf{Document-Level Rules. }
Documents that match any of following rules will be discarded.

\begin{itemize}
\item If document does not contain between 50 and 100,000 words;
\item If mean word length is outside the range of 3 to 10 characters;
\item If the ratio of the number of hash symbols and ellipses to the total word count exceeds 0.1;
\item If more than 90\% of sentences in the document start with bullet point;
\item If more than 30\% of sentences in the document end with an ellipsis;
\item If more than 20\% of words in the document doesn't contain alphabetic characters;
\item If a document does not contain at least two stop words, such as \texttt{the}, \texttt{be}, \texttt{to}, \texttt{of}, \texttt{and}, \texttt{that}, \texttt{have}, and \texttt{with}.
\end{itemize}

\item \textbf{Sentence-Level Rules. }
Sentences that match any of following rules will be removed, if over 5\% words of the document is removed by sentence filters the whole document is discarded.

\begin{itemize}
\item If over 60\% of letters in the sentence is uppercase;
\item If it only contains numerical characters;
\item If it is a counter (match regex \texttt{\^\textbackslash{}d+\textbackslash{}s+[a-zA-Z]+\$}, e.g., \texttt{3 likes});
\item If it only contains one word;
\item If matches of following regex could be found in the sentence
\begin{itemize}
    \item \textasciicircum{}sign-in
    \item read more...\$
    \item items in card
\end{itemize}
\end{itemize}

\end{itemize}

\paragraph{Model-based Filtering.} Rule-based filtering can only identify fixed patterns within the text and cannot assess its quality in terms of syntax, content, or overall coherence.
We employed a transformer-based classifier for data filtering, following a process similar to FineWeb-edu~\cite{lozhkov2024fineweb-edu}. We first sampled data pre-filtered by rule-based methods and had GPT-4 annotate 2,000 samples for quality, evenly split between positive and negative. We then trained stablelm-1.6b on this annotated data to filter the entire dataset. The filtering criteria focused on grammar, logic, and most importantly, knowledge—defined as data that enhances the model's real-world understanding. This approach ensures the model is trained on high-quality, informative data, improving its performance on downstream tasks.

\paragraph{Deduplication.}
The presence of substantial duplicate content on the web necessitates deduplication to reduce the required training steps and increase data diversity~\cite{carlini2022quantifying, lee2021deduplicating}. Many studies have demonstrated that deduplication can enhance model performance. 
For deduplication, we utilized MinHash~\cite{broder1997resemblance}, a popular and efficient technique for detecting duplicate content in large datasets. MinHash approximates the Jaccard similarity between sets, making it suitable for identifying near-duplicate texts without the need for exhaustive pairwise comparisons, which can be computationally prohibitive for massive datasets.

We employ the MinHash-LSH algorithm to extend the deduplication scope across all snapshots. 
We generate 117 64-bit MinHash signatures (effectively 62-bit due to the limitations of the Meson prime) over 5-gram sequences of documents. These signatures are divided into 9 bands, each containing 13 values, to achieve an estimated 80\% Jaccard similarity when an LSH collision occurs.

To reduce computational complexity and enable a broader processing range and faster iteration speed, we opted against using larger MinHash parameters or connected-graph-based deduplication methods.

\subsection{\ours{}-Code}
\label{sec:data_code}
Code is a significant domain for LLMs, particularly in areas such as code generation, code summarization, and code refinement. Leveraging capabilities across different areas, various downstream applications have been developed to improve code quality, accelerate code writing, and fix code bugs. For example, GitHub Copilot\footnote{\url{https://github.com/features/copilot}} has been adopted by more than 50,000 organizations and is favored by approximately 55\% of developers.

To advance the development of code-related capabilities, several initiatives have focused on creating comprehensive code datasets. One notable example is StarCoder~\cite{li2023starcoder}, which aggregates code data from the BigCode community. However, these datasets typically consist solely of source code and lack the contextual information or explanations accompanying the code snippets. Unlike previous efforts, our goal is to construct an interleaved code dataset using \ours{} that includes both source code and the interleaved text providing context or explanations for the code to enhance understanding and usability.

The original 89 snapshots of Common Crawl contain over 200 billion HTML documents, making efficient processing of these documents a challenging task. We designed the following steps, utilizing the filtering and extraction modules of \ours{}. First, we employ the filtering module to exclude pages that are not related to code. Subsequently, we use the extraction module to convert the HTML pages into plain text required for training.

Upon the completion of the aforementioned steps, we have obtained approximately 114.9 million documents, encompassing a total of 250.2 billion tokens, as detailed in Table~\ref{tab:dataset}. The detailed steps are as follows.

\subsubsection{Filtering}
Since code snippets in HTML pages are often enclosed in special tags (e.g., \texttt{<code>}), these tags can be used to quickly determine whether a page contains code snippets. Therefore, \ours{}-Code uses WARC files as input and employs code tags to filter out pages that do not contain code snippets. Additionally, during the processing of HTML pages, we observed that they often contain a significant amount of noise (e.g., hidden nodes) that could interfere with subsequent content extraction. To address this, we implemented element filtering to remove invisible elements. Moreover, Code snippets sometimes appear within a single element, such as inline or single-line code snippets. Other times, they span multiple elements that share a common parent, such as multi-line code snippets. In these cases, we use rules to determine the overall scope of the code, filter out smaller code elements, and merge the code snippets into a single cohesive unit. The specifics of each step are detailed below.

\begin{itemize}
\item \textbf{Document Filtering.}
To expedite the processing, we implement a pre-filtering operation on the raw HTML content using keyword filters. Specifically, (1) only HTML documents that include the \texttt{<code} or \texttt{<pre} keyword are passed to the next step; (2) To reduce the interweaving situation of two code snippets under the code diff or blame function, we exclude documents whose URLs include \texttt{blame.php} or \texttt{diff.php}.

\item \textbf{HTML Element Filtering.}
There are several ways to hide HTML elements, such as using \texttt{aria-hidden="true"}, \texttt{display:none}, and \texttt{visibility:hidden}. Some of these are set in HTML, while others are set via JavaScript or CSS. To prevent duplicated or invisible content from remaining after text extraction, we remove all elements with these properties in the HTML.

\item \textbf{Refined Code Filtering.}
Code snippets in HTML documents can be represented in various ways. Sometimes, a snippet is enclosed in a single \texttt{<code>} element, such as with inline and single-line code snippets. Other times, it spans several \texttt{<code>} elements that share a common parent, as seen with multi-line code snippets. To filter HTML documents for code snippets and identify the root element of each filtered code snippet, we follow these steps:

\begin{itemize}
\item Traverse each element in the DOM tree of the HTML document and identify all elements of type \texttt{<code>}.
\item If the parent of each \texttt{<code>} element is either \texttt{<pre>} or \texttt{<tbody>}, treat its parent as a candidate root element for containing the code snippet. Otherwise, the \texttt{<code>} element itself is the candidate root element.
\item Apply a regular expression to the extracted text from each candidate root element to determine whether it includes a code snippet. Specifically, the candidate root element is considered to contain a code snippet if it matches one of the four conditions in the designed regular expression: keywords of programming languages, code indicators, function calls, and variable assignments.
\item If a candidate element is identified as containing a code snippet, its type is changed to \texttt{<code-encode>}. Consequently, the entire HTML document is detected as containing code knowledge.
\item To avoid the interweaving situation of code lines and line numbers in one code snippet, we remove line numbers by identifying and deleting consecutive single-line or alternating-line increasing numbers within each code snippet.
\item To merge adjacent code snippets that should belong to one code snippet, we detect and combine them by locating a closing code tag immediately followed by an opening code tag.
\end{itemize}

\end{itemize}

\subsubsection{Extraction}
In this step, we convert the WARC HTML pages to WET plain text format by adopting the WEATGenerator function in the \texttt{ia-hadoop-tools} library~\cite{CC23}, which is the WARC-to-WET conversion method officially used by Common Crawl. We do not use Trafilatura to extract the main content from the HTML because Trafilatura's recall is relatively low, and it is likely to exclude code snippets from the main content.

\subsection{\ours{}-Math}
\label{sec:data_math}
The ability to reason mathematically is crucial for LLMs. Several studies have aimed to enhance this capability. For instance, the PaLM~\cite{chowdhery2023palm} was fine-tuned on billions of tokens from mathematical documents sourced from arXiv and the web, while OpenWebMath ~\cite{paster2023openwebmath} was trained on 14.7 billion tokens of mathematical webpages from Common Crawl. Both models have shown significant improvements in solving problems that require quantitative reasoning.

Mathematical formulas or code snippets on web pages typically have corresponding HTML tags, making the logic for constructing these two types of data quite similar. There are only slight differences in the processing steps. By making appropriate modifications, we utilized the HTML content from Common Crawl's WARC files to filter pages containing mathematical content on a large scale.

Additionally, during the construction of \ours{}-Math, we discovered a substantial number of mathematical formulas in ASCII format. These formulas exist as plain text on web pages and cannot be identified through HTML tags. Therefore, for these mathematical pages, we designed a new processing workflow specifically to extract these formulas.

Finally, we compiled a comprehensive \ours{}-Math totaling 15.9 billion tokens, as illustrated in Table~\ref{tab:math_data_scale}. This dataset encompasses mathematical formulas along with their corresponding context, thereby enhancing the model's ability to understand mathematics. The following sections will describe the construction processes for these two different types of mathematical data.

\begin{table}[htbp]
\centering
\begin{tabular}{lccc}
\hline
\bf \ours{}-Math & \bf Tokens (B) & \bf Pages (M) & \bf Tokens per Page \\
\hline
HTML-Math  & 11.1 & 2.7 & 4,041.2 \\
ASCII-Math & 4.8 & 3.7 & 1,297.9 \\
\hline
Total & 15.9 & 6.4 & 2,460.7 \\
\hline
\end{tabular}
\caption{\ours{}-Math from Common Crawl.}
\label{tab:math_data_scale}
\end{table}

\subsubsection{HTML-Math}

\paragraph{Filtering.}
The filtering process encompasses document filtering, HTML element filtering, and fine-grained filtering. The workflow is similar to that used for \ours{}-Code. Document filtering removes pages that do not contain mathematical formulas by using a coarse heuristic based on HTML tags typically associated with mathematics. However, content wrapped in mathematical tags is not necessarily a valid formula; it could merely be a few mathematical characters or incorrect expressions. Therefore, more refined mathematical formula filtering rules are required to accurately determine whether a page genuinely contains mathematical formulas. The specifics of each step are detailed below.

\begin{itemize}
\item \textbf{Document Filtering. }
We use keywords such as \texttt{<math}, \texttt{<annotation}, \texttt{="math}, \texttt{athjax}, \texttt{math-container}, \texttt{class="tex"}, \texttt{tex.cgi}, \texttt{latex.php}, \texttt{katex.min.css}, \texttt{\textbackslash frac}, and \texttt{codecogs} to pre-filter the HTML documents associated with mathematics.

\item \textbf{Refined Math Filtering. }
To verify the syntactical correctness of formulas, it is essential first to identify their positions for extraction. Consequently, this process is divided into two steps: formula localization and formula syntax verification.

\begin{itemize}
\item To identify mathematical formulas in HTML documents, we use HTML tags and related attributes. For instance, we utilize tags such as \texttt{<script>} and \texttt{<math>}. For \LaTeX~formulas, we use specific prefixes like \texttt{<script type="math/tex"}, \texttt{<script type="math/latex"}, \texttt{<script type="math/asciimath"}, \texttt{<span class="math-formula"}, and \texttt{<annotation encoding="application/x-tex"} to detect them. For MathML formulas, we apply \texttt{<math} and \texttt{<script type="math/mml"} to identify them.

\item To ensure the grammatical correctness of each mathematical formula, we apply \texttt{LatexWalker} from the \texttt{pylatexenc} library~\cite{phfaist23} to check the grammatical correctness of each formula, filtering out documents with errors.

\end{itemize}
\end{itemize}

Another potential issue is the possibility of duplicate representations of formulas. For instance, a \LaTeX~annotation, typically represented as \texttt{<annotation encoding="application/x-tex"}, is included as child node within a MathML formula represented as \texttt{<math>}. This could result in multiple duplicates of the same formula. Therefore, we need to ensure that each mathematical formula is represented only once, eliminating any duplicate representations, such as formula text along with formula annotations, formula images, and alternative text for images.

\paragraph{Extraction.} 
The Extraction module follows the same processing procedure as 
\ours{}-Code, which will not be reiterated here. 

\subsubsection{ASCII-Math}
During the processing, we observed that many mathematical formulas were directly represented in ASCII math format.
These formulas were not enclosed within any math-related HTML tags but existed as plain text within the pages. This type of data is also highly valuable. For this type of data, we employed the following pipeline.

\paragraph{Extraction.}
Since the mathematical formulas are no longer encapsulated within HTML tags, we can convert HTML pages into plain text format in advance to facilitate subsequent filtering. Unlike the text extraction process described for HTML tags, all HTML documents in this step are converted into text documents using the Trafilatura, excluding menu, header, and footer content, resulting in cleaner text.

\paragraph{Filtering.}
To efficiently extract pages containing mathematical formulas from a vast collection of raw pages, we employ a two-stage filtering approach. Initially, we use rule-based heuristics to quickly and broadly exclude pages without mathematical content. Subsequently, we apply a model for finer-grained filtering on the remaining pages, generating the final dataset.

\begin{itemize}
\item \textbf{Rule-based Filtering.}
We categorize the keywords into two groups: \LaTeX~symbols and non-\LaTeX~symbols. For \LaTeX~symbols, such as \texttt{\textbackslash frac}, \texttt{\textbackslash mu}, \texttt{\textbackslash dot}, \texttt{\textbackslash log}, and \texttt{\textbackslash eq}, we have collected over 3,000 keywords in total. For non-\LaTeX~symbols, such as \texttt{sqrt}, \texttt{sum}, \texttt{log}, \texttt{+}, \texttt{*}, and \texttt{\$}, we have gathered more than 20 keywords. Based on these keywords, we filter out text documents that contain fewer than five of them. At this stage, only about 0.1\% of the text documents pass the filter and proceed to the next stage for further processing.

\item \textbf{Model-based Filtering.} 
After the pre-filtering stage by keywords, we further filter out the text documents by employing a self-trained classification model. Firstly, we construct the training dataset by sampling 100k text documents from the pages previously identified as containing mathematical formulas using HTML tags, and also randomly selecting 100k text documents that exclude the math text documents as negative samples, and totally 200k samples. Secondly, we train an n-gram model using fasttext~\cite{joulin2016fasttext} to balance efficacy and efficiency. During the filtering stage, we apply this self-trained model to further filter the text documents, using a threshold of 0.5.

\end{itemize}

\subsection{\ours{}-QA}
\label{sec:data_qa}
QA datasets often contain a wealth of specific knowledge and are highly valuable for advancing understanding in various fields. To answer these questions, LLMs need to have a comprehensive and accurate understanding of the world, requiring critical thinking, analysis, and discussion. We have created two types of QA datasets: open question answering and multiple-choice question answering.

For the open question answering data, there are various types, such as short answer, fill-in-the-blank, multiple choice, and true-or-false questions. These open questions are often interspersed with other text content in the document, which may include reading passages, explanations, or other informative content. Therefore, the contextual information along with the open questions is essential for a language model to effectively understand and learn.

For the multiple-choice question answering data, during the construction of open question answering data, we observed a significant number of high-quality, valuable multiple-choice questions within the Common Crawl. A considerable proportion of these multiple-choice questions include the stem, options, answers, and explanations. The presence of answers and explanations imbues this data with a high density of knowledge, facilitating a deeper understanding of the real world for the model. Therefore, we further processed the open question answering data to obtain multiple-choice question answering data.

\begin{table}[htbp]
\centering
\begin{tabular}{lccc}
\hline
\bf \ours{}-QA & \bf Tokens (B) & \bf Pages (M) & \bf Tokens per Page \\
\hline
Open Question Aswering & 51.3 & 42.5 & 1,206.8 \\
Multi-choice Question Aswering  & 0.1 & 1.6 & 63.8 \\
\hline
Total & 51.4 & 44.1 & 1165.5 \\
\hline
\end{tabular}
\caption{\ours{}-QA from Common Crawl.}
\label{tab:data_qa}
\end{table}

The final dataset of \ours{}-QA consists of 44.1 million documents and 51.4 billion tokens, as shown in Table~\ref{tab:data_qa}. The following sections will describe the data construction process in detail.

\subsubsection{Open Question Aswering}

In processing pages that contain open questions, we utilized the extraction and filtering modules. The extraction module is responsible for extracting main content from the pages, while the filtering module is tasked with identifying and selecting the pages that contain open questions. The specific steps are as follows.

\paragraph{Extraction.}
The first step is to extract the main content of the original pages. Due to the low recall when extracting main content from WARC files using Trafilatura, which often results in missing parts of the main content, we use WET files as input for creating the \ours{}-OpenQA. To ensure the completeness of the extracted content, we employed CCNet's paragraph-level exact deduplication method. This extraction procedure is consistent with the one described earlier and can be referenced in Section~\ref{sec:general_data_extraction}.

\paragraph{Filtering.}
To quickly and efficiently identify pages containing open-domain questions, our filtering module employs a two-stage approach: rule-based filtering and model-based filtering. Initially, we use rule-based filtering to identify common patterns of open-domain questions, thereby discarding pages that are unlikely to contain such questions. This step significantly reduces the number of pages. Subsequently, we apply model-based filtering to the remaining pages to detect the presence of open-domain questions.

\begin{itemize}
\item \textbf{Rule-based Filtering.} 
Since the answers contain the knowledge for each question, We filter text documents to ensure they include both questions and answers by designing two sets of patterns for identifying each. For the question pattern, 
we use keywords such as ``what'', ``where'', ``why'', ``when'', ``who'', ``whose'', ``how'', ``q\&a'', ``q \& a'', ``q:'', ``que:'', ``question:'', ``quiz:'', ``exam:'', ``examination:'', ``probe:'', ``request:'', ``challenge:'', ``test:'', ``query:'', and ``survey:''. The document must contain at least one of these keywords. 
For the answer pattern, we utilize keywords such as ``q\&a'', ``q \& a'', ``a:'', ``ans:'', ``answer:'', ``solution:'', ``reply:'', ``response:'', ``result:'', ``outcome:'', ``explanation:'', ``conclusion:'', ``finding:'', ``assertion:'', ``statement:'', and ``clarification:''. 
The document must also contain at least one of these keywords. Only text documents that match both patterns are retained for further filtering. 

\item \textbf{Model-based Filtering.} 
Rule-based filtering serves only as an initial pre-processing step. To further refine the text documents and ensure they contain open questions, we employ a self-trained classification model. Specifically, (1) we use GPT-3.5-turbo as an annotator to label whether a page contains open questions and to categorize them (e.g., short answer, fill-in-the-blank, multiple choice, etc.), and then constructed a training set of 200k labeled samples, with an equal split between positive and negative samples. Among the positive samples, we ensured a balanced mix of different types of open-domain questions. (2) We then trained an n-gram-based classification model using fastText~\cite{joulin2016fasttext} to determine whether a page contains open-domain questions. (3) Finally, we used the trained model to filter data at scale, thereby obtaining a large dataset of open-domain questions.
\end{itemize}

\subsubsection{Multi-choice Question Aswering}

The open question aswering data contains pages with various types of open questions, which need to be further filtered to obtain pages containing multiple-choice questions. For multiple-choice questions, we only extract content related to the questions, such as the question stems, answers, and explanations, and finally standardize this content into a uniform format. 

We utilize two modules in the \ours{}: Filtering and Extraction. The Filtering module identifies pages containing multiple-choice questions, while the Extraction module extracts the multiple-choice questions from these pages.

\paragraph{Filtering.}
Among the components of multiple-choice questions, the list of options is more distinctly characterized and thus easier to locate. Consequently, to efficiently filter out pages containing multiple-choice questions from the vast number of web pages, we employed the following two rules:

\begin{itemize}
\item The serial number is typically in one of the following formats: \texttt{a, b, c, d, \ldots}, \texttt{1, 2, 3, 4, \ldots}, and \texttt{i, ii, iii, iv, \ldots}.

\item The delimiter can be one of the following: \texttt{.}, \texttt{-}, \texttt{)}, \texttt{$>$}, \texttt{]}, and \texttt{\textbackslash space}.
\end{itemize}

Additionally, we ensure that multiple-choice questions have corresponding answers by retaining only those text documents that contain one of the following keywords: \texttt{answer:}, \texttt{solution:}, \texttt{reply:}, \texttt{response:}, \texttt{ans:}, \texttt{a:}, and \texttt{r:}.
By applying these rules, we can quickly filter and identify a large number of pages that potentially contain multiple choice questions.

\paragraph{Extraction.}
We aim to obtain a clean dataset containing only multiple choice questions. Therefore, from the pages identified by the filtering module as potentially containing multiple choice questions, we further locate and extract the questions from the pages. It is important to note that multiple choice questions without answers are useless for model training, so we only extract those that include answers, explanations for the answers are optional.

\begin{itemize}
\item The serial number is typically in one of the following formats: \texttt{a, b, c, d, \ldots}, \texttt{1, 2, 3, 4, \ldots}, and \texttt{i, ii, iii, iv, \ldots}.

\item The delimiter can be one of the following: \texttt{.}, \texttt{-}, \texttt{)}, \texttt{$>$}, \texttt{]}, and \texttt{\textbackslash space}.
\end{itemize}

To locate the multiple choice questions on a page, we use different rules to identify each component: the stem, the options, the answer, and the explanation. A multiple choice question is retained only if it contains at least the stem, options, and answer.  we employed the following rules:

\begin{itemize}
\item \textbf{Choice List Identification:}
Locate the candidate of the choice list for each multiple choice question using the rules concluded in the filtering stage.

\item \textbf{Stem Identification:}
The stem usually precedes the choice list, so the textline before the options is selected as the question stem.

\item \textbf{Answer Identification:}
If the text line following each candidate in the choice list starts with any of the following keywords: \texttt{answer:}, \texttt{solution:}, \texttt{reply:}, \texttt{response:}, \texttt{ans:}, \texttt{a:}, or \texttt{r:}, we treat this candidate as a genuine choice list. The text line following the choice list is added as the answer, while the text line preceding it is considered the question.

\item \textbf{Explanation Identification:}
If the text line following the answer starts with \texttt{explanation:}, it is added as the explanation component as well.

\end{itemize}

Given the various formats of multiple choice questions on the web—for instance, the answer might precede the stem—the order of the components is not fixed. Therefore, after extracting the multiple choice questions, we standardize each component to ensure uniformity. For example, we arrange each multiple choice question in the order of question, option list, answer, and explanation. We prefix the answer with \texttt{Answer:} and, if an explanation is present, we prefix it with \texttt{Explanation:}. The steps are as follows:

\begin{itemize}
\item \textbf{Question Formatting:}
Remove any serial number and delimiter at the start of the question.
\item \textbf{Choice List Formatting:}
Convert the serial numbers of the choice list to \texttt{A, B, C, D, ...} in order and use \texttt{.} as the delimiter between the serial number and choice.
\item \textbf{Answer and Explanation Formatting:}
Precede the answer with \texttt{Answer:} and the explanation, if it exists, with \texttt{Explanation:}.
\item \textbf{Component Arrangement:}
Arrange each multiple choice question in the order of the question, choice list, answer, and explanation.
\end{itemize}

%% file: chapters/experiments.tex
\section{Experiments}

\subsection{Settings}
\textbf{Model Setting.} 
For evaluations on general domain datasets, we trained models with 1.3 billion parameters on 50 billion tokens. We utilized the same architecture as the LLaMA model, incorporating SwiGLU~\cite{shazeer2020glu}, RoPE~\cite{su2024roformer}, and RMS~\cite{zhang2019root}. For evaluations on the domain-specific dataset, We utilized the same architecture as the StableLM-2-1.6B\cite{bellagente2024stable} model. As described in Table~\ref{tbl:hyperparam:Model:pt:opt}.

\begin{table}[ht]
\centering
\centering
\begin{tabular}{lcc}
\toprule
\textbf{Hyperparameters} & \bf General Domain & \bf Domain-specific \\ \midrule
Layers   &       24 &  32 \\
Embed Dim     &      2,048  &  2,560 \\
FFN Dim & 5,760  & 6,912 \\
Attn Head & 32  & 32 \\
Vocab Size & 32k  & 50k \\
\bottomrule
\end{tabular}
\caption{Hyperparameters of model}
\label{tbl:hyperparam:Model:pt:opt}
\end{table}

\textbf{Training Setting.} For evaluations on general domain datasets, We follow the GPT-3 paper to set the training parameters. For evaluations on the domain-specific dataset, The learning rate is 2e-5, total training steps is 10k, batch size is 16, max token size is 2k. Further details can be found in Table~\ref{tbl:hyperparam:training:pt:opt}.

\begin{table}[ht]
\centering
\centering
\begin{tabular}{lcc}
\toprule
\textbf{Hyperparameters} & \bf General Domain & \bf Domain-Specific\\ \midrule
Training steps   &       300,000 & 10,000 \\
Warmup steps     &       1,000 & 1,000 \\
Optimizer & \multicolumn{2}{c}{AdamW} \\
Learning rate & 2e-4  & 2e-5 \\
Learning rate decay & \multicolumn{2}{c}{Linear} \\
Adam $\beta$ & \multicolumn{2}{c}{(0.9, 0.95)} \\
Weight decay & \multicolumn{2}{c}{0.1} \\
Dropout & \multicolumn{2}{c}{0} \\
\midrule
Batch size of text         &      512   & 16 \\
\#Token per Batch & 1M  & 32k \\
\bottomrule
\end{tabular}
\vspace{0.2cm}
\caption{Hyperparameters of training}
\label{tbl:hyperparam:training:pt:opt}
\end{table}

\subsection{Evaluation Datasets}
To comprehensively evaluate our proposed datasets, we employed a two-pronged approach, utilizing different benchmarks for general domain and domain-specific datasets.

\begin{table}[ht]
    \centering
    \begin{tabular}{llc}
    \toprule
        \textbf{Tasks} & \textbf{Type}  \\
        \midrule
         HellaSwag \cite{zellers2019hellaswag} & Common Sense  \\
         Winogrande \cite{sakaguchi2019winogrande} & Common Sense  \\ 
         PIQA \cite{bisk2020piqa} & Common Sense \\ 
         ARC-E \cite{clark2018arc} & Question Answering  \\ 
         ARC-C \cite{clark2018arc} & Question Answering  \\ 
         OpenBookQA \cite{mihaylov2press2021train018openbookqa} & Question Answering \\
        \bottomrule
    \end{tabular}
    \caption{Evaluation tasks of general domain data}
    \label{tab:evaluation_tasks_of_general}
\end{table}

For the general domain dataset, we leveraged the lm-evaluation-harness pipeline\footnote{\url{https://github.com/EleutherAI/lm-evaluation-harness}}, which includes a variety of tasks that test reading comprehension, common sense reasoning, and question answering abilities. The benchmarks used for this evaluation are listed in Table \ref{tab:evaluation_tasks_of_general}.

\begin{table}[ht]
    \centering
    \begin{tabular}{llc}
    \toprule
        \textbf{Tasks} & \textbf{Type}  \\
        \midrule
         HumanEval \cite{chen2021codex} & Code Synthesis \\
         MBPP \cite{austin2021program} & Code Synthesis  \\
         GSM8k \cite{cobbe2021gsm8k} & Mathematics \\
         Minerva Math \cite{dyer2022minerva} & Mathematics \\
         MMLU \cite{hendryckstest2021, hendrycks2021ethics} & Common Sense  \\
         Winogrande \cite{sakaguchi2019winogrande} & Common Sense \\ 
         ARC-C \cite{clark2018arc} & Question Answering \\ 
         ARC-E \cite{clark2018arc} & Question Answering \\
         OpenBookQA \cite{mihaylov2press2021train018openbookqa} & Question Answering\\
        \bottomrule
    \end{tabular}
    \caption{Evaluation tasks of domain-specific data}
    \label{tab:tab:evaluation_tasks_of_specific}
\end{table}

For the domain-specific datasets, we selected benchmarks that are tailored to assess the performance of models in code generation, mathematical reasoning, and question-answering tasks. These benchmarks are detailed in Table \ref{tab:tab:evaluation_tasks_of_specific}.

\subsection{\ours{}-Web}

\begin{table*}[t!]
  \centering
  \small
  \setlength{\tabcolsep}{3pt}
  \begin{tabular}{lrcccccc}
  \toprule
\bf Datasets    & \bf ARC-C  & \bf ARC-E  & \bf HellaSwag & \bf OpenBookQA & \bf PIQA & \bf Winogrande & \bf AVERAGE\\
  \midrule
  RedPajama    & 0.2270 & 0.4386 & 0.3171 & 0.1900 & 0.5968 & \textbf{0.5296} & 0.3832 \\
  FineWeb      & 0.1928 & 0.4428 & 0.3506 & 0.1740 & 0.6681 & 0.5288 & 0.3929 \\
  RefinedWeb   & 0.2125 & 0.4369 & 0.3380 & 0.2100 & 0.6491 & 0.5264 & 0.3955   \\
  DCLM         & 0.2159 & 0.4848 & 0.3614 & 0.1760 & 0.6615 & 0.5082 & 0.4013 \\
  FineWeb-Edu  & \textbf{0.2722} & \textbf{0.5648} & 0.3637 & 0.1940 & 0.6676 & 0.5051 & 0.4279  \\
  \ours{}-Web  & 0.2662 & 0.5181 & \textbf{0.3722} & \textbf{0.2340} & \textbf{0.6795} & 0.5162 & \textbf{0.4310}  \\
  \bottomrule
  \end{tabular}
  \caption{
  Comparison of evaluation tasks across open source datasets.
  \label{tab:results_general}
  }
\end{table*}

We evaluated the performance of our general domain dataset, \ours{}-Web, across a variety of common sense reasoning tasks and compared it with several other well-known open-source datasets. The results of these evaluations are summarized in Table \ref{tab:results_general}. The results demonstrate that 
\ours{}-Web performs exceptionally well in most tasks, with particularly strong performance in the ARC-E, HellaSwag, OpenBookQA, and PIQA tasks. Specifically, \ours{}-Web achieved the highest scores in HellaSwag (0.3722), OpenBookQA (0.2340), and PIQA (0.6795), indicating that it is highly effective in capturing the nuances required for these tasks.

The comprehensive evaluation across these tasks reveals that 
\ours{}-Web is a highly competitive dataset for general pre-training. It surpasses the other datasets in most common sense reasoning tasks. This indicates that our approach of leveraging Common Crawl to extract a diverse and high-quality dataset is effective in enhancing the capabilities of LLMs.

\subsection{\ours{}-Code}

\begin{table}[htbp]
\centering
\setlength{\tabcolsep}{3pt}
\begin{tabular}{lcccc}
\hline
{\bf Datasets} & \makecell{\bf HumanEval \\ pass@1} & \makecell{\bf HumanEval \\ pass@10} & \makecell{\bf MBPP \\ pass@1} & \makecell{\bf MBPP \\ pass@10}\\
\hline
\ours{}-Web  & 0.0125 & 0.0168 & 0.0751  & 0.1566 \\
~~~~ + \ours{}-Code & \textbf{0.0555} & \textbf{0.1035} & \textbf{0.1311} & \textbf{0.2458} \\
\hline
\end{tabular}
\caption{Evaluation of \ours{}-Code. ``+'' indicates the combination of the current dataset with the previous one. In the experiments, the weight of the \ours{}-Web data is set to 0.8, and the weight of the \ours{}-Code data is set to 0.2.
}
\label{tab:result_webcode}
\end{table}

The results in Table \ref{tab:result_webcode} demonstrate the significant performance improvement achieved by incorporating the \ours{}-Code dataset into the \ours{}-Web dataset. For HumanEval pass@1, the combination yields a score of 0.0555, a substantial increase from the 0.0125 achieved by \ours{}-Web alone. Similarly, HumanEval pass@10 improves from 0.0168 to 0.1035. This indicates a marked improvement in the model's ability to generate correct solutions within the first attempt as well as within ten attempts. A similar trend is observed in the MBPP dataset, where the pass rates also exhibit considerable advancements. The pass@1 metric rises from 0.0751 to 0.1311, and the pass@10 metric increases from 0.1566 to 0.2458.

These experimental results indicate that \ours{}-Code can serve as a valuable supplement to LLM pre-training datasets in the domain of code. This integration lays a strong foundation for enhancing the code generation capabilities of LLMs in downstream tasks.

\subsection{\ours{}-Math}

\begin{table}[t]
\centering
\begin{tabular}{lcccccccc}
\toprule
\bf Datasets & \bf GSM8k & \bf MATH \\
\midrule
OpenWebMath \cite{paster2023openwebmath}  & 3.2503 & 3.1288\\
\ours{}-Math & \textbf{3.1125} & \textbf{3.0557} \\
\bottomrule
\end{tabular}
\caption{Evaluation of \ours{}-Math and perplexity measurement on various mathematics benchmarks. Both OpenWebMath and \ours{}-Math were trained from scratch.}
\vspace{-1.5em}
\label{tab:result_webmath}
\end{table}

Table \ref{tab:result_webmath} presents a comparative evaluation of \ours{}-Math and OpenWebMath on two mathematics benchmarks, GSM8k and MATH. The evaluation metric used is perplexity, which scales more smoothly than accuracies.

From the results, it is evident that \ours{}-Math outperforms OpenWebMath on both datasets. Specifically, \ours{}-Math achieves a perplexity of 3.1125 on the GSM8k dataset, compared to OpenWebMath's perplexity of 3.2503. This represents a significant improvement, suggesting that \ours{}-Math enables faster convergence and enhances modeling capability for the types of problems presented in GSM8k. Similarly, on the MATH dataset, \ours{}-Math demonstrates a perplexity of 3.0557, whereas OpenWebMath records a perplexity of 3.1288. This further underscores the effectiveness of \ours{}-Math in handling diverse mathematical problems.

\ours{}-Math consistently achieves lower perplexity scores across both evaluated datasets, highlighting its potential to enhance model performance in the domain of mathematics. It can serve as a valuable supplement to pre-training datasets in this field.

\subsection{\ours{}-OpenQA}

\begin{table}[htbp]
\centering
\setlength{\tabcolsep}{1pt}
\begin{tabular}{lcccccc}
\toprule
\bf Model & \makecell{\bf MMLU} & \makecell{\bf ARC-Challenge} & \makecell{\bf ARC Easy} & \makecell{\bf Openbookqa} & \makecell{\bf Winogrande} & \bf AVERAGE\\
\midrule
StableLM-2-1.6B         & 0.3135 & 0.3481 & \textbf{0.6860} & 0.2780 & 0.6354 & 0.4522 \\
~~~~ + FALN v2          & 0.3525 & 0.3601 & 0.6406 & \textbf{0.2860} & 0.6125 & 0.4503\\
~~~~ + Open Orca        & 0.3569 & 0.3089 & 0.5821 & 0.2660 & 0.5675 & 0.4163\\
~~~~ + \ours{}-QA & \textbf{0.4582} & \textbf{0.3643} & 0.6839 & 0.2760 & \textbf{0.6377} & \textbf{0.4840} \\
\bottomrule
\end{tabular}
\caption{Evaluation of \ours{}-QA: \ours{}-QA comprises two components—open question answering and multiple-choice question answering, with training weights set to 0.8 and 0.2, respectively.
}
\label{tab:result_web_oq}
\end{table}

Table \ref{tab:result_web_oq} presents the performance evaluation of various models across multiple benchmark datasets, including MMLU, ARC Challenge, ARC Easy, OpenbookQA, and Winogrande. The baseline model, StableLM-2-1.6B, demonstrates robust performance with an average accuracy of 0.4522. After fine-tuning with FALN v2, the model shows slight improvements on the MMLU and ARC Challenge tasks, although the overall average performance slightly decreases from 0.4522 to 0.4503. Fine-tuning with Open Orca maintains competitive performance on the MMLU task but leads to significant drops in scores on the ARC Easy and OpenbookQA tasks, from 0.6860 to 0.5821 and from 0.2780 to 0.2660, respectively. The average performance decreases to 0.4163, indicating that this enhancement may not generalize well across different tasks.

The proposed \ours{}-QA dataset achieves the highest scores on most datasets, with an overall average performance significantly improving to 0.4840. Notably, it shows a 14.47\% improvement on the MMLU dataset, highlighting the effectiveness of \ours{}-QA in enhancing the model's question-answering capabilities.

%% file: chapters/related_works.tex
\section{Related Work}

\subsection{General Domain Data Pipelines} 
Since the advent of BERT~\cite{devlin2018bert}, there has been a growing trend to leverage large-scale data from Common Crawl for model training. CCNet~\cite{wenzek2019ccnet} extracted a substantial amount of high-quality bilingual text from it to build machine translation models. T5~\cite{raffel2020exploring} employed a series of heuristic rules to extract text and create the C4 dataset. With the rapid development of LLMs, recent efforts have increasingly focused on constructing larger and higher-quality datasets from Common Crawl to support the training of these models. The Pile~\cite{gao2020pile} used jusText~\cite{endredy2013more} to extract text from Common Crawl, resulting in Pile-CC. LLaMA~\cite{touvron2023llama} utilized the CCNet pipeline with modifications to generate a vast amount of pre-training data, though this data was not made publicly available. Subsequently, RedPajama~\cite{together2023redpajama} reproduced the training data used in LLaMA and open-sourced the data. Aiming for even higher data quality, RedPajama v2~\cite{together2023redpajama} introduced 46 quality signals to describe data characteristics from various dimensions. RefinedWeb~\cite{penedo2023refinedweb} employed content extraction tools to extract main content from HTML pages provided by Common Crawl, yielding cleaner and higher-quality text, though only a small portion of this data was open-sourced. In response, FineWeb~\cite{penedo2024fineweb} reproduced RefinedWeb and open-sourced the data, while also constructing a new filter to exclude educational content, resulting in the higher-quality pre-training dataset FineWeb-edu. DCLM~\cite{li2024datacomp} extracted a large volume of text from Common Crawl and developed a custom filter to obtain a substantial amount of instruction-formatted data, significantly enhancing data quality. We also introduce \ours{}-Web, a high-quality dataset derived from our \ours{} pipeline, which features simpler processing steps, more data, and improved quality.

\subsection{Domain Specific Data Pipelines} 
LLMs are currently being widely utilized to address a variety of problems. Among the most prevalent applications are their roles as powerful assistants for writing and editing code, as well as for solving mathematical problems. Additionally, LLMs are frequently engaged in interactive QA formats, making QA capabilities a critical aspect of their evaluation. Given the extensive range of potential applications for LLMs, this paper focuses on advancements in code, math, and QA, which are also key components of \ours{}.

\textbf{Code.} Recent advancements in code generation have been significantly driven by Code LLMs. Prominent models such as CodeGen ~\cite{nijkamp2022codegen}, ERNIE-Code ~\cite{chai2022ernie}, StarCoder ~\cite{lozhkov2024starcoder}, CodeT5+ ~\cite{wang2021codet5}, CodeLLaMa ~\cite{roziere2023code}, and Deepseek-Coder ~\cite{guo2024deepseek} focus on enhancing coding capabilities during the pre-training stage. Given the substantial data requirements for pre-training, GitHub has emerged as a crucial resource ~\cite{li2022competition, laurenccon2022bigscience, nijkamp2022conversational, fried2022incoder}. For instance, The Stack utilized GHArchive to download 137.36 million repositories from GitHub. After filtering, cleaning, license detection, and deduplication, a training set of 1450.75 GB of data spanning 30 popular programming languages was obtained ~\cite{kocetkov2022stack}. Subsequently, The Stack v2 expanded its data sources to include other high-quality open datasets, such as GitHub issues, pull requests, Kaggle and Jupyter notebooks, code documentation, and other natural language datasets related to math, coding, and reasoning. This expansion resulted in a dataset four times larger than the initial StarCoder dataset. Additionally, models like Code Alpaca ~\cite{codealpaca}, WizardCoder ~\cite{luo2023wizardcoder}, OctoPack ~\cite{muennighoff2023octopack}, and Magicoder ~\cite{wei2023magicoder} focus on enhancing coding capabilities during the instruction tuning stage. This stage typically involves constructing high-quality instruction datasets to improve model performance.

\textbf{Math.}
The application of LLMs in solving mathematical problems has seen significant advances, driven by the necessity to manage complex computations and deliver precise solutions. Models such as GPT-3 ~\cite{brown2020language} and Minerva ~\cite{lewkowycz2022solving} have been specifically fine-tuned for mathematical problem-solving tasks. These models utilize extensive datasets comprising mathematical texts, problem sets, and solutions to enhance their capabilities. For instance, the OpenWebMath dataset ~\cite{paster2023openwebmath}, which includes 14.7 billion tokens of mathematical web pages from Common Crawl, demonstrates the potential of training models on large-scale mathematical data. The training process often involves step-by-step problem-solving, aiding in the understanding and generation of solutions for complex mathematical queries. Furthermore, incorporating symbolic computation and formal methods into LLMs has enhanced their ability to handle algebraic manipulations, calculus, and other advanced mathematical concepts. The integration of these techniques ensures that the models provide not only accurate answers but also human-readable solutions that are easy to follow and verify.

\textbf{Question and Answer.} Interactive Question and Answer (QA) capabilities are a cornerstone of LLM applications. Models like BERT ~\cite{devlin2018bert}, T5 ~\cite{raffel2020exploring}, and GPT-3 ~\cite{brown2020language} have set benchmarks in the QA domain by excelling in tasks such as reading comprehension, open-domain QA, and conversational agents. The performance of these models is often evaluated on standardized datasets like SQuAD ~\cite{rajpurkar2016squad}, TriviaQA ~\cite{joshi2017triviaqa}, and Natural Questions ~\cite{kwiatkowski2019natural}. Additionally, the Massive Multitask Language Understanding (MMLU) dataset ~\cite{hendryckstest2021} has emerged as a critical benchmark for evaluating the broad knowledge and reasoning abilities of LLMs across diverse subjects, ranging from elementary mathematics to advanced science and humanities. Recent models, such as ChatGPT ~\cite{OpenAI2023ChatGPT} and InstructGPT ~\cite{ouyang2022training}, build on these foundations by incorporating reinforcement learning from human feedback (RLHF) to improve their interactive and conversational abilities. Specialized models like QA-GNN ~\cite{yasunaga2021qa} and DrQA ~\cite{chen2017reading} enhance QA performance by integrating graph neural networks and retrieval-based approaches, respectively. These advancements enable LLMs to provide more accurate, context-aware, and informative answers, making them indispensable tools for customer support, educational tutoring, and information retrieval systems. The continuous refinement of QA datasets and the development of hybrid models that combine retrieval and generative approaches are driving the next generation of QA systems.

%% file: chapters/conclusion.tex
\section{Conclusion and Future Work}

This paper introduces \ours{}, a comprehensive data pipeline designed to create specialized large-scale datasets by leveraging the vast and diverse data from Common Crawl. The \ours{} pipeline is built around {Extraction} and {Filtering} modules. These modules can be flexibly combined to efficiently mine data from various domains. We used \ours{} to construct large-scale datasets, totaling 3.48 trillion tokens, for \ours{}-Web, \ours{}-Code, \ours{}-Math, and \ours{}-QA. The experimental results demonstrate the effectiveness of \ours{}. This dataset not only showcases the pipeline's ability to extract specific types of data but also significantly surpasses existing open-source datasets in terms of quality and scale. We have detailed the dataset construction process to ensure reproducibility and transparency, providing a valuable resource for developing competitive LLMs. This makes it possible to create high-quality, domain-specific datasets at scale. The datasets and code developed through \ours{} will be open-sourced to foster further development and collaborative efforts within the research community.

For future work, several promising directions could enhance \ours{}'s capabilities. First, integrating more advanced filtering techniques, such as leveraging multimodal signals (e.g., visual or contextual cues) and sophisticated machine learning models, could refine data quality and improve domain diversity. Second, extending the pipeline to support multilingual datasets and incorporating multimodal content, such as images, audio, and videos, would enable the development of LLMs that excel in cross-lingual and multimodal tasks. Lastly, implementing real-time data updates, including automated mechanisms to incorporate fresh and relevant web data while maintaining stringent quality standards, would ensure the datasets remain current and applicable to evolving use cases. Together, these efforts will unlock the full potential of large-scale web data, advancing the creation of specialized and high-performance language models across diverse domains.

%% file: chapters/appendix.tex
\appendix

\section{Data Samples}

\input{chapters/cases}

%% file: chapters/cases.tex
\tcbset{
    breakable, enhanced,   
    left*=10pt, right*=10pt,
    top=0pt, bottom=0pt,
    colback=white!10!white,
    colframe=black!75!black,
    fonttitle=\bfseries\large,
    subtitle style={boxrule=0pt,colback=gray!50!white},
    before=\vspace{-1em}, %
    after=\vspace{1em} %
}

\lstset{language=python, breaklines=true}
\begin{tcolorbox}[title=\ours{} - Web]
\scriptsize
\setlength{\tabcolsep}{0.1mm}{
\begin{xltabular}{\linewidth}{X}

Water is the biggest necessity for a tree. It uses it for photosynthesis and loses the water as it travels through the tree and eventually evaporates. With the hot Dallas heat and limited amounts of rain, it’s common to see trees begin to wilt and lose life because of their inability to withstand the heat without the proper water intake.
\newline \newline 
Many people don’t even think about watering their trees. They assume that their home irrigation system will water the tree just perfectly or that it rains just enough to keep the trees healthy for the time being, but that’s not the case. Home sprinkler systems do not provide the tree with enough water and do not place the water where it needs to go in order to be efficiently absorbed by the roots.
\newline \newline 
There’s a technique to watering trees. The water has to be able to travel beneath the surface surrounding the base of the tree and if too much water has been accumulated, the tree could contract a disease and die. Absorbing water through its bark is not something that can provide a tree with the right amount of nutrients to survive. The water has to enter from the base to be absorbed straight into the roots.
\newline \newline
When you water your tree follow these next few steps:
\newline \newline
Organic mulch
\newline
Using an all-natural organic mulch can significantly help the tree’s water intake. Surround the tree with this mulch ad begin to water it. The mulch can absorb water easier than a chemically-filled version, allowing the water to pass through its surface and into the ground, reaching the roots.
\newline \newline
Do not directly water its base
\newline
Avoid watering the tree directly at its base. If you do, the water will not be able to seep into the ground at a fast enough rate and the stagnant water can cause your tree to obtain a disease. Begin watering about three feet from the base and continue moving away from the tree in the areas that the foliage covers. This will provide consistent watering that travels deep into the ground. Get optimum results when you use the aforementioned organic mulch.
\newline \newline
Water at night
\newline
We understand that you want to have the best landscaping Dallas has to offer at your home, and your trees are what pushes it to the next level. So when you water them, make sure that you are doing so at night. Watering them during the day with the high temperatures and strong sun proves pointless. The sun can easily evaporate most of the water as it tries to seep into the ground. Avoid wasting water by watering your trees during the evening hours when the sun isn’t so strong.
\newline \newline
Remember to water your tree at least twice a week due to the lack of rain and high temperatures. Contact your local Scapes, Inc. to receive professional tree care services in the Dallas area.

\\
\midrule
Carbon Build Up (And Why You Need To Know About It)
\newline \newline
The Issue
\newline \newline
Carbon build up is a common problem with petrol engines in generators, both large and small. When carbon builds up in an engine, airflow is restricted and causes problems. Carbon build up increases when a generator is used below sufficient load and at a constant RPM. This can damage the engine and reduce its life and operating efficiency.
\newline \newline
The Details
\newline \newline
Carbon build up is caused by running an engine at low loads. This can also happen when an engine is left idling, as a'standby' unit. Cylinders and pistons rely on higher pressures to operate effectively, which isn't happening when running on low loads. This results in poor combustion, leading to soot formation and unburnt fuel - which is bad for your engine!
\newline \newline
The Symptoms
\newline \newline
- The engine not running smoothly
\newline
- The engine vibrating or shaking
\newline
- Jerking or surging
\newline
- Smoke (never a good sign!)
\newline \newline
How To Avoid It
\newline \newline
Carbon build up can be minimised by using your generator at higher loads more often - just like a body needs exercise to work its best, an engine needs to be run at higher loads to continue operating at its best.
\newline \newline
However, carbon build up cannot be completely avoided. It is highly important to get your generator serviced about every 12 months (dependent on how often you use it). Macfarlane Generators have highly qualified and experience service technicians, who know the in's and out's of carbon build up, and other common problems (such as stale fuel). By getting a service every 12 months, we will be able to predict and prevent issues before they become issues!
\\

\end{xltabular}}
\end{tcolorbox}

\newpage

\begin{tcolorbox}[title=\ours{} - Code]
\scriptsize
\setlength{\tabcolsep}{0.1mm}{
\begin{xltabular}{\linewidth}{X}
Borrowing from the example on the Matplotlib documentation page and slightly modifying the code,

<code-encode><metadata lang=Python prob=0.93 />

import numpy as np

from mpl\_toolkits.mplot3d import Axes3D

import matplotlib.pyplot as plt

def randrange(n, vmin, vmax):

\hspace{2em}return (vmax-vmin)*np.random.rand(n) + vmin
    
fig = plt.figure()

ax = fig.add\_subplot(111, projection='3d')

n = 100

for c, m, zl, zh in [('r', 'o', -50, -25), ('b', '\^', -30, -5)]:

\hspace{2em}xs = randrange(n, 23, 32)
    
\hspace{2em}ys = randrange(n, 0, 100)
    
\hspace{2em}zs = randrange(n, zl, zh)
    
\hspace{2em}cs = randrange(n, 0, 100)
    
\hspace{2em}ax.scatter(xs, ys, zs, c=cs, marker=m)
    
ax.set\_xlabel('X Label')

ax.set\_ylabel('Y Label')

ax.set\_zlabel('Z Label')

plt.show()

</code-encode>

Will give a 3D scatter plot with different colors for each point (random colors in this example). What's the correct way to add a colorbar to the figure, since adding in plt.colorbar() or ax.colorbar() doesn't seem to work. 

\\
\midrule

The LV2 documentation generation tools use RDFLib. It is probably the most popular RDF interface for Python, though does much more than just parse Turtle. It is a good choice if performance is not an issue, but is unfortunately really slow.

If you need to actually instantiate and use plugins, you probably want to use an existing LV2 implementation. As Steve mentioned, Lilv is for this. It is not limited to any static default location, but will look in all the locations in LV2\_PATH. You can set this environment variable to whatever you want before calling Lilv and it will only look in those locations. Alternatively, if you want to specifically load just one bundle at a time, there is a function for that: lilv\_world\_load\_bundle().

There are SWIG-based Python bindings included with Lilv, but they stop short of actually allowing you to process data. However there is a project to wrap Lilv that allows processing of audio using scipy arrays: http://pyslv2.sourceforge.net/ (despite the name they are indeed Lilv bindings and not bindings for its predecessor SLV2)

That said, if you only need to get static information from the Turtle files, involving C libraries is probably more trouble than it is worth. One of the big advantages of using standard data files is ease of use with existing tools. To get the number of ports on a plugin, you simply need to count the number of triples that match the pattern (plugin, lv2:port, *). Here is an example Python script that prints the number of ports of a plugin, given the file to read and the plugin URI as command line arguments:

<code-encode><metadata lang=Python prob=0.84 />

\#!/usr/bin/env python

import rdflib

import sys

lv2 = rdflib.Namespace('http://lv2plug.in/ns/lv2core\#')

path = sys.argv[1]

plugin = rdflib.URIRef(sys.argv[2])

model = rdflib.ConjunctiveGraph()

model.parse(path, format='n3')

num\_ports = 0

for i in model.triples(plugin, lv2.port, None]):

\hspace{2em}num\_ports += 1

print('\%s has \%u ports' \% (plugin, num\_ports))

</code-encode>

\end{xltabular}}
\end{tcolorbox}

\newpage

\begin{tcolorbox}[title=\ours{} - Math]
\scriptsize
\setlength{\tabcolsep}{0.1mm}{
\begin{xltabular}{\linewidth}{X}
The Maxwell-Boltzmann Distribution describes the distribution of the speed of particles in an idealized gas. It plays an important role in understanding the kinetic energy distribution of electrons and ions as well as in characterizing a particular gaseous substance. In classical physics, it was thought that the molecules of ideal gases bounced around with arbitrary velocities, never interacting with one another. However, scientists later discovered that an ideal gas is better modeled by taking intermolecular interactions into account.  
  
The Maxwell-Boltzmann distribution is a probability density function for the speed \(v\) of molecules with mass \(m\) in a gas at absolute temperature \(T\). It is given by the following equation:  
  
\[   
f\left(v\right) = 4 \pi \left(\frac{m}{2 \pi k T}\right)^{3/2} v^2 e^{-\frac{m v^2}{2 k T}}  
\]  
  
where \(k\) is Boltzmann's constant, \(1.38 \times 10^{-23}\) J/K. The probability that the molecule will have any speed between \(v_1\) and \(v_2\) is given by:  
  
\[   
P\left(v_1 < v < v_2\right) = \int_{v_1}^{v_2} f\left(v\right) dv  
\]  
  
The mean speed, \(\overline{v}\), can be calculated as follows:  
  
\[   
\overline{v} = \int_{0}^{\infty} v f\left(v\right) dv = \sqrt{\frac{8 k T}{\pi m}}  
\]  
  
The mean speed is always higher than the most probable speed as a result of the skewness of the distribution. Adjust the sliders to see how temperature and mass affect the Maxwell-Boltzmann Distribution.  
  
What is the average speed of nitrogen gas molecules (each with a mass of \(4.65 \times 10^{-26}\) kg) when they are exposed to a temperature of 25 °C (298.15 K)?  
  
You can use the formula given in the previous section to determine the average speed:  
  
\[   
\overline{v} = \sqrt{\frac{8 \left(1.38 \times 10^{-23} \text{ J/K}\right) \left(298.15 \text{ K}\right)}{\pi \left(4.65 \times 10^{-26} \text{ kg}\right)}}  
\]  
  
\[   
\overline{v} = 471 \frac{m}{s}  
\]  
  
Therefore, the average speed is 471 m/s.
\\

\midrule

Use sigma notation to write the right Riemann sum for \( \textstyle f(x)=x+2 \) on the interval \( \textstyle [0,6] \) with \( \textstyle n=60 \). **DO NOT EVALUATE** the right Riemann sum.  
  
The right Riemann sum for \( \textstyle f(x) \) with the partition \( \textstyle P=\big\{ [x_{0},x_{1}],[x_{1},x_{2}],\dots ,[x_{n-1},x_{n}] \big\} \) is given by \( \displaystyle \sum_{i=1}^{n}f(x_{i})(x_{i}-x_{i-1}) \). Given \( \displaystyle n=60 \) and the given interval for the Riemann sum is \( \displaystyle [0,6] \). This implies that each interval in the partition has the size \( \displaystyle \frac{1}{10}=\frac{6}{60} \), so we will use the partition \( \textstyle P=\big\{ [0,\frac{1}{10}],[\frac{1}{10},\frac{2}{10}],\dots ,[6-\frac{1}{10},6] \big\} \).  
  
Then plugging \( \textstyle f(x)=x+2 \) and \( \textstyle n=60 \) into the formula in the hint, we see that the right Riemann sum is precisely  
  
\[   
\displaystyle \sum_{i=1}^{60} \frac{f(i/10)}{10} = \sum_{i=1}^{60} \left(\frac{i}{10}+2\right) \frac{1}{10}.  
\]  
  
**Answer**: The right Riemann sum is given by   
  
\[   
\displaystyle \sum_{i=1}^{60} \left(\frac{i}{10}+2\right) \frac{1}{10}.
\] 

\\

\end{xltabular}
}
\end{tcolorbox}

\newpage

\begin{tcolorbox}[title=\ours{} - OpenQA]
\scriptsize
\setlength{\tabcolsep}{0.1mm}{
\begin{xltabular}{\linewidth}{
    X
    }
\textbf{Open Question Aswering}

I've been searching everywhere for the driver for my Forsa Fs 3120N printer for Windows XP, but I just can't seem to find it. I've checked the manufacturer's website and various driver download sites, but no luck. Can anyone please provide me with a direct link to download the driver for my printer? I would greatly appreciate it. Thank you in advance.
\newline
\newline
Detective Site Admin Below you can download forsa fs 3120n win xp driver for Windows. 
\newline
File name: forsa\_fs-3120n-win-xp.exe 
\newline
Version: 3.822 
\newline
File size: 14.71 MB 
\newline
Upload source: torrents 
\newline
Antivirus software passed: Kaspersky v 2.9 
\newline
Download Driver Forsa Fs 3120N Win Xp (click above to download) 
\newline
Device Specification: 
\newline
Device Model: Forsa Fs 3120N 
\newline
Supported Operating System: Windows XP 
\newline
Input: Download Driver for Forsa Fs 3120N 
\newline
Output: N/A (since it only states to download the driver) 
\newline
\newline
Troubleshooting Questions and Answers: 
\newline
1. Q: Why is my Forsa Fs 3120N not working after installing the downloaded driver? 
\newline
A: Make sure you have installed the correct driver version compatible with your Windows XP operating system. Restart your computer after installation and check if the device is recognized by the system. 
\newline
2. Q: Can I use the Forsa Fs 3120N on other operating systems like Windows 10 or macOS? 
\newline
A: The Forsa Fs 3120N is specifically designed to be compatible with Windows XP. It may not work properly or may require different drivers for other operating systems. It is recommended to check with the manufacturer or official support channels for compatibility and driver availability on different OS. \newline
3. Q: How can I download the driver for the Forsa Fs 3120N? 
\newline
A: To download the driver for the Forsa Fs 3120N, search for the official Forsa website or support page. Look for the specific driver for Windows XP for your device model. Once found, follow the instructions provided on the website to download and install the driver on your computer.
\\
\midrule
\textbf{Open Question Aswering}

Question: What is a collection?
\newline
Answer: A collection is a group of publications, all of which have a specific character and can be described by a certain set of attributes (qualities), used to search the publications. Some
attributes can be found in various collections, while others are specific to
only one collection.
\newline
\newline
Question: How to search by collections?
\newline
Answer: Included on the main page of the Digital Library of University of Wroclaw is a whole list of collections and subcollections. Clicking on the name of a collection will allow to display the whole list of publications within the collection (link: Publications list).
\newline
\newline
Question: What is an exhibition?
\newline
Answer: An exhibition is a set of publications belonging to the same subject and being presented in a specific order. Documents included in an exhibition may come from various collections.
\newline
\newline
Question: What is the general search (simple and advanced) and how to use it?
\newline
Answer: Each publication consists of a description (comprising a set of attributes, such as author, title, editor, keywords etc.) and a content. In order to find a requested publication it is
necessary for a user to create a search expression using terms (in the simple search) or terms and the Boolean operators (in the advanced search). In the simple search the terms appearing in the search expression are searched in a content of a publication, in a description of a publication (it is possible to narrow the search scope to only one element of the description, e.g. an author) or in both a content and a description of the publication (when the search scope is set to everywhere, which is a default value). In the advanced search the terms appearing in the search expression are searched in a content and (or) in a description of a publication (it is possible to narrow the search scope to only
one element of the description), using the Boolean operators. The general search may be conducted on the whole digital library (all collections are searched) or on a chosen collection (all publications within the collection are searched). The second option allows to shorten the search time and limits the number of results.
\newline
\newline
Question: What is the index search and how to use it?
\newline
Answer: Index search, unlike the general search (simple and advanced) limits the search only to descriptions of the publications. There are indexes of titles, authors and keywords available in the index search.
\newline
Question: What is the use of a user account?
\newline
Answer: A user account gives the user access to
the additional functionality, e.g. an e-mail newsletter containing a list of recently added publications. To create a user account it is necessary to have an e-mail account, to which an e-mail with an account activation link will be sent. In order to activate the account the user
has to click on the received activation link.
\newline
\newline
Question: Is it possible to download a publication without opening it?
\newline
Answer: Yes. To download a publication in ZIP format the “floppy disk” icon can be used. The same effect may be obtained by using the Publication window (in the left column) and clicking on the link
Entire publication as ZIP.
\newline
\newline
Question: In what ways does the option “Use synonyms” help?
\newline
Answer: Using synonyms expands the search by including terms with the same or similar meanings, which
are listed in a dictionary of synonyms, created by editors of the digital library. This option enables to obtain results indicating publications which contain synonyms of requested search terms.
\\
\midrule
\textbf{Multi-choice Question Aswering}

Which diagram in UML emphasizes the time-ordering of messages?\newline
A. Activity\newline
B. Sequence\newline
C. Collaboration\newline
D. Class\newline
Answer:A\newline
Explanation:This diagram is a model describing how groups of objects collaborate in some behavior over time. 
\\

\midrule
\textbf{Multi-choice Question Aswering}

Pick the odd one out.\newline
A. File transfer\newline
B. File download\newline
C. E-mail\newline
D. Interactive games\newline
Answer:D\newline
Explanation: File transfer, File download, and Email are services provided by the application layer and there are message and data-oriented. 

\end{xltabular}}
\end{tcolorbox}

\newpage